\documentclass[10pt,twocolumn,letterpaper]{article}

\usepackage{iccv}
\usepackage{times}
\usepackage{epsfig}
\usepackage{graphicx}
\usepackage{amsmath}
\usepackage{amssymb}
\usepackage{tablefootnote}
\usepackage[labelformat=empty]{subfig}
\usepackage{multirow}
\usepackage{float}
\usepackage{xcolor}

\usepackage[breaklinks=true,bookmarks=false]{hyperref}

\iccvfinalcopy 

\def\iccvPaperID{****} 

\ificcvfinal\fi

\begin{document}

\title{Ref-DVGO: Reflection-Aware Direct Voxel Grid Optimization for an Improved Quality-Efficiency Trade-Off in Reflective Scene Reconstruction}

\author{Georgios Kouros\\
KU Leuven\\
{\tt\small georgios.kouros@esat.kuleuven.be}
\and
Minye Wu\\
KU Leuven\\
{\tt\small minye.wu@esat.kuleuven.be}
\and
Shubham Shrivastava\\
Ford Motor Company\\
{\tt\small sshriva5@ford.com}
\and
Sushruth Nagesh\\
Ford Motor Company\\
{\tt\small snagesh1@ford.com}
\and
Punarjay Chakravarty\\
(Former) Ford Motor Company\\
{\tt\small punarjay@gmail.com}
\and
Tinne Tuytelaars\\
KU Leuven\\
{\tt\small tinne.tuytelaars@esat.kuleuven.be}
}

\maketitle
\ificcvfinal\thispagestyle{empty}\fi

\begin{abstract}
Neural Radiance Fields (NeRFs) have revolutionized the field of novel view synthesis, demonstrating remarkable performance.
However, the modeling and rendering of reflective objects remain challenging problems.
Recent methods have shown significant improvements over the baselines in handling reflective scenes, albeit at the expense of efficiency.
In this work, we aim to strike a balance between efficiency and quality.
To this end, we investigate an implicit-explicit approach based on conventional volume rendering to enhance the reconstruction quality and accelerate the training and rendering processes. 
We adopt an efficient density-based grid representation and reparameterize the reflected radiance in our pipeline. 
Our proposed reflection-aware approach achieves a competitive quality-efficiency trade-off compared to competing methods. 
Based on our experimental results, we propose and discuss hypotheses regarding the factors influencing the results of density-based methods for reconstructing reflective objects. The source code is available at: \url{https://github.com/gkouros/ref-dvgo}
\end{abstract}

\section{Introduction}

%
Neural Radiance Fields (NeRF) \cite{mildenhall2020nerf} rekindled research interest in novel view synthesis by producing high-quality photorealistic renderings thus facilitating the development of countless applications and new computer vision tasks.
However, NeRF fails to produce photorealistic renderings of scenes with highly reflective effects, which are fairly common in any environment, be it domestic, urban, or industrial.

The main challenge of neural rendering on reflective objects lies in the fact that the color of a single point on a given surface can differ considerably depending on the viewing angle, illumination, scene layout, as well as material properties of that surface.
This introduces a shape-appearance ambiguity that erroneously guides a regular NeRF model to wrongly optimize the geometry of the scene in such a way that allows the reconstruction of the training images while failing to produce plausible novel views.
The result of this ambiguity is semi-transparent surfaces with foggy artifacts embedded inside the volume of the reflective object. 
Ref-NeRF \cite{verbin2022refnerf} addresses this issue by proposing a reparameterization of the outgoing radiance as a function of the reflection direction which facilitates more efficient interpolation of reflections and specular highlights. 
However, Ref-NeRF requires considerable training time and resources due to its two large MLPs that are queried thousands of times per iteration. 
Subsequent works \cite{ge2023ref,fan2023factoredneus,Liang2023ENVIDRID} employ the signed distance function (SDF) rather than density-based representations, and along with other reflection-aware design choices, they  achieve both reduced training time and either improved rendering quality and/or a more refined geometry. 
However, they still require at least an order of magnitude more training time and resources compared to more efficient methods \cite{SunSC22dvgo,mueller2022instant}, while at the same time, they do not support (semi-)transparent surfaces. 

Improving the training or rendering speed of neural rendering models has been widely researched and relevant methods have concentrated on either incorporating priors \cite{roessle2022depthpriorsnerf,yu2020pixelnerf}, using multi-resolution hash encodings \cite{mueller2022instant} or replacing the implicit MLP-based representations with explicit voxelized structures \cite{SunSC22dvgo,yu_and_fridovichkeil2021plenoxels,Chen2022ECCV}.
DVGO \cite{SunSC22dvgo}, in particular, employs a hybrid implicit-explicit representation and achieves training times over two orders of magnitude faster than fully implicit methods with competitive rendering quality which, however, degrades drastically in the case of reflective scenes.

In this work, we investigate the quality-efficiency trade-off in combining a fast hybrid implicit-explicit representation, i.e. DVGO, and the reflection-aware architecture of Ref-NeRF with a view to achieving high-quality photorealistic novel view synthesis of challenging reflective scenes in tens of minutes rather than hours.
To achieve this, first, we use the proposed architecture of Ref-NeRF with the reflection direction reparameterization and simply replace the spatial MLP with six distinct voxel grids, one for each of its outputs, as depicted in Fig. \ref{fig:methodology:model-architecture}.
%
%
We evaluate our approach on the shiny blender dataset \cite{verbin2022refnerf} that contains six synthetic reflective objects, as well as on a custom single scene car dataset \footnote{The dataset was generated using the blender model from \url{https://www.lightmap.co.uk/learning/blender-tutorial-03/}} with varying levels of reflectivity. Transparent and semi-transparent surfaces/objects are currently not examined. In our experiments, we demonstrate improved reconstruction quality compared to the baselines (NeRF, iNGP, DVGO), although, still below the performance of Ref-NeRF, but with a considerably faster training and rendering speed and with four times less GPU memory requirements. We also present our hypotheses on the reasons our method outperforms DVGO but not Ref-NeRF.

\textbf{}\section{Neural Rendering of Reflective Objects with a Hybrid Implicit-Explicit Representation}


\begin{figure}[t]
    \centering
    \includegraphics[width=\linewidth]{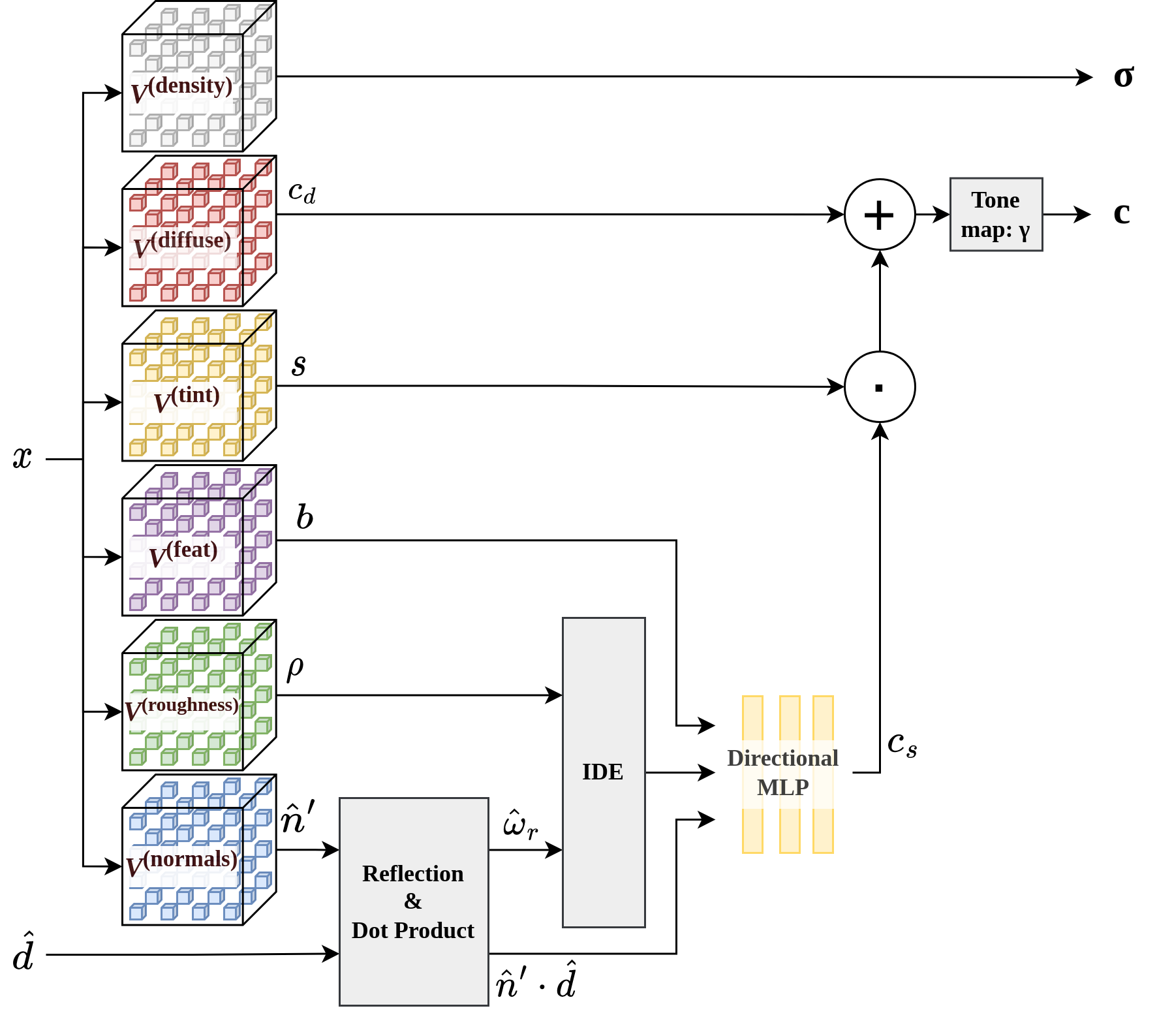}
    \vspace{-0.7cm}
    \caption{Our proposed Ref-DVGO architecture aims to leverage the rendering quality of Ref-NeRF \cite{verbin2022refnerf} on reflective objects with the training and rendering speed of the hybrid implicit-explicit representation of DVGO \cite{SunSC22dvgo}.}
    \label{fig:methodology:model-architecture}
\end{figure}

Our approach, similar to Ref-NeRF, utilizes a reparameterization of the outgoing radiance as a function of the reflection direction rather than the view direction, and expresses the outgoing radiance at every point in space by combining the incoming radiance, diffuse color, roughness, and specular tint properties of the scene. 
As demonstrated in \cite{verbin2022refnerf}, these updates facilitate smoother interpolation of specular highlights and reflections. 
One drawback in the architecture of Ref-NeRF lies in the fact that it utilizes large MLPs that slow down training and rendering.
To alleviate this effect, we replaced the spatial MLP of Ref-NeRF with six voxel grids that represent view-independent properties of the scene, namely density $\sigma$, diffuse color $c_d$, specular tint $s$, appearance/bottleneck features $b$, roughness $\rho$, and predicted normals $\hat n'$, respectively, as shown in Fig.~\ref{fig:methodology:model-architecture}.\\[-0.2cm]

\noindent \textbf{Optimization.} The resulting model, which we call Ref-DVGO, is optimized using a combination of the losses from \cite{verbin2022refnerf,SunSC22dvgo,sun2022improved}. Similar to all NeRF works, the optimization is led by the photometric loss $\mathcal{L}_{ph}$ \cite{mildenhall2020nerf}, which supervises rendered and alpha-composed ray colors $\hat C(r)$ with the ground truth pixel colors $C(r)$ of the training images.
%
%
Besides supervising the rendered and alpha-composited ray colors, we also directly supervise the colors $c_i$ of each of the $K$ sampled points along a ray $r$ using the per-point RGB Loss $\mathcal{L}_{pp}$ from \cite{SunSC22dvgo}.
%
%
In addition, the background entropy loss $\mathcal{L}_{bg}$ of DVGO \cite{SunSC22dvgo} regularizes the rendered background probability to reduce uncertainty between foreground and background. 
%
%
Furthermore, to improve the estimation of the surface normals of the scene, we employ the predicted normals penalty $\mathcal{R}_p$ and the normal orientation regularization $\mathcal{R}_o$ \cite{verbin2022refnerf}. The former supervises the predicted normals $\hat n_i'$ with the normals $\hat n_i$ estimated through the density gradient resulting in overall smoother normals. The latter penalizes back-facing normals to concentrate all normals on the surface of the object and prevent semi-transparent surfaces and embedded or floating artifacts.
%
%
%
%
%
%
The final loss function is calculated as the weighted sum of the aforementioned losses and regularizers as \\[-0.35cm]
\begin{equation*}
\begin{split}
    \mathcal{L} = w_{ph} \mathcal{L}_{ph} + w_{pp} \mathcal{L}_{pp} + w_{bg} \mathcal{L}_{bg} + w_{p} \mathcal{R}_p + w_o \mathcal{R}_o,
    \label{eq:methodology:totalloss}
\end{split}
\end{equation*}
\noindent where $w_{ph}$, $w_{pt\_rgb}$, $w_{bg}$, $w_{p}$, and $w_{o}$ are the weights of the aforementioned losses and regularizers.
Last but not least, as a means to reduce geometry noise and unnecessary sharpness, as well as to improve spatial continuity, we apply total variation regularization $\mathcal{R}_{TV}$ \cite{rudin1994tv,sun2022improved,yu_and_fridovichkeil2021plenoxels} on the gradients of all six voxel grids, rather than just density.\\[-0.2cm]


\noindent \textbf{Training Stages.} Our method is built on top of DVGO \cite{SunSC22dvgo}, so we naturally employ a coarse-to-fine training scheme, which tends to be faster than using proposal networks and self-distillation as in \cite{barron2021mipnerf,barron2022mipnerf360,mildenhall2020nerf}.
First, a coarse model with only view-independent components (density, diffuse color) is trained for a limited number of iterations (i.e. 5000) to learn a coarse geometry and refine the bounding box of the scene. 
Then, with the view-dependent components enabled, the fine stage optimizes the fine model by efficiently sampling the coarse model for points in occupied space.
This informed sampling greatly reduces the number of redundant queries to the directional MLP resulting in considerably faster training and rendering times.
Last but not least, the fine stage is applied in a coarse-to-fine manner using progressive scaling (PGS) of the voxel grid.
%
%
This ensures that the model learns the low-frequency information of the scene first before the high-frequency details and view-dependent effects, which are harder to optimize.
%

\section{Results}

\begin{figure*}
\begin{center}
\begin{tabular}{cccccc}
\subfloat{\includegraphics[width=0.16\linewidth]{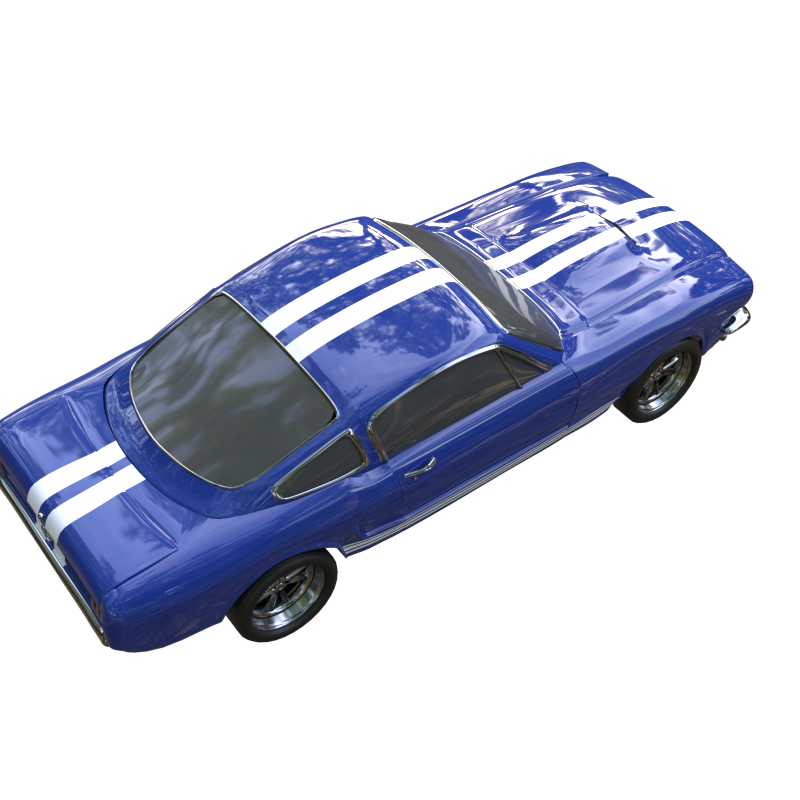}}\hspace*{\fill}
\subfloat{\includegraphics[width=0.16\linewidth]{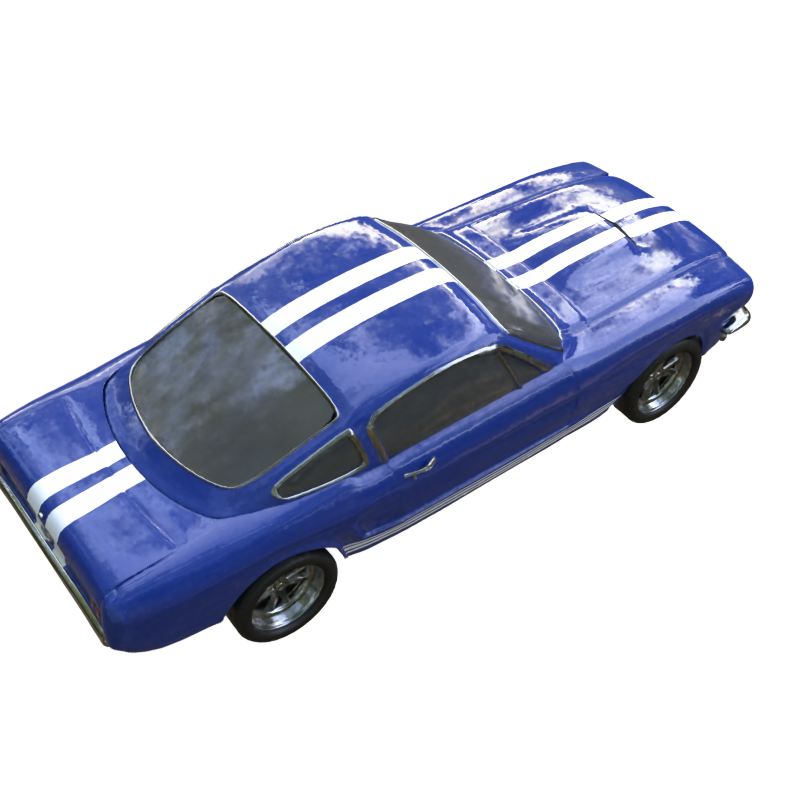}}\put(-28,20){\small{29.43}}\hspace*{\fill}
\subfloat{\includegraphics[width=0.16\linewidth]{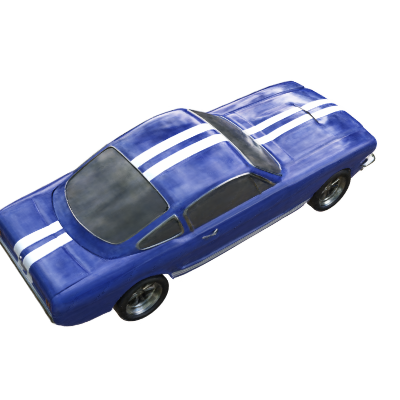}}\put(-26,20){\small{26.79}}\hspace*{\fill}
\subfloat{\includegraphics[width=0.16\linewidth]{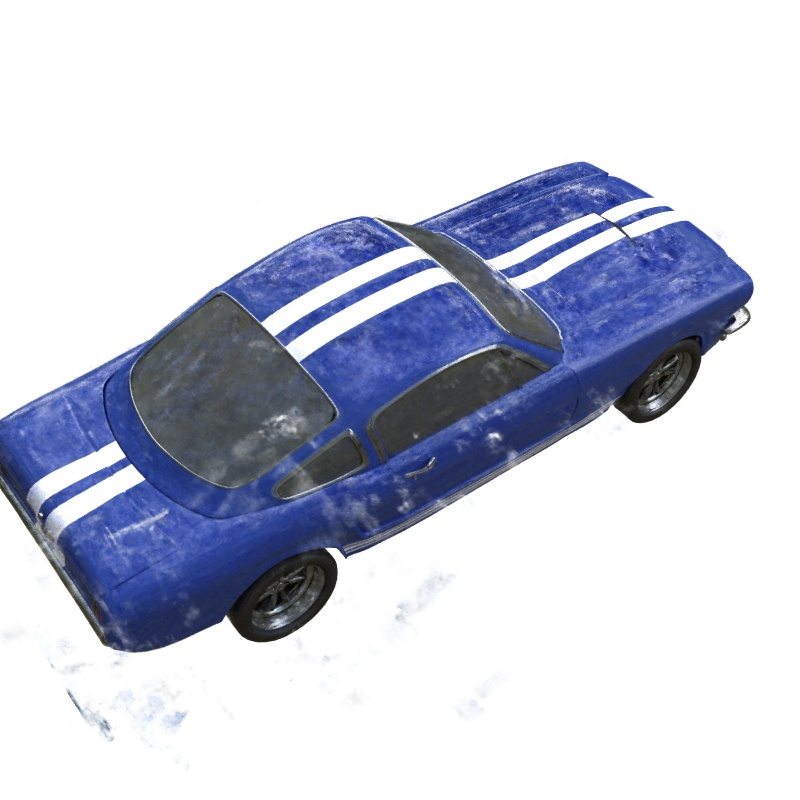}}\put(-24,20){\small{24.99}}\hspace*{\fill}
\subfloat{\includegraphics[width=0.16\linewidth]{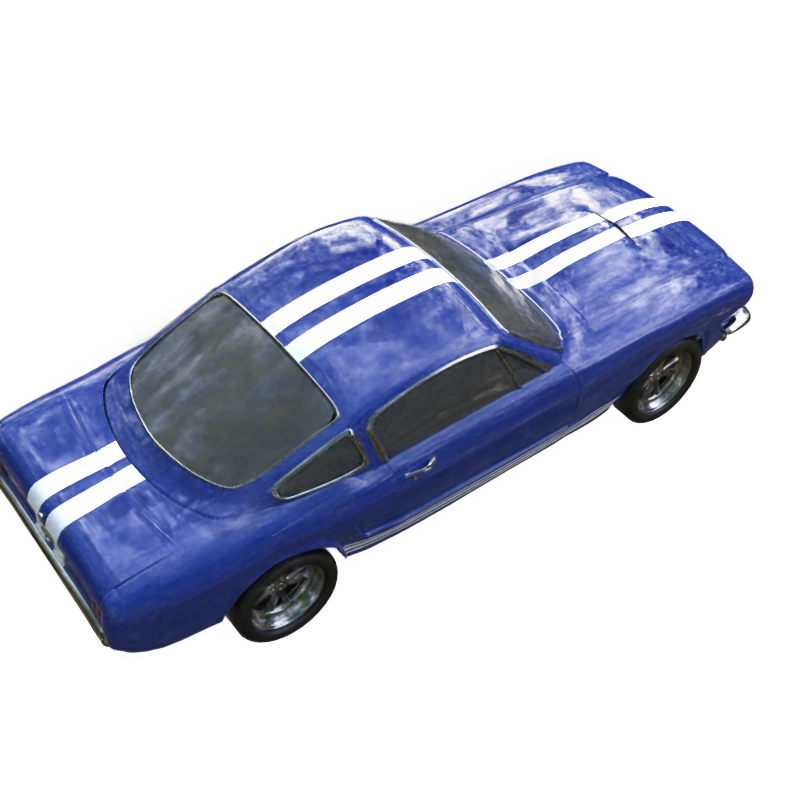}}\put(-22,20){\small{26.99}}\hspace*{\fill}
\subfloat{\includegraphics[width=0.16\linewidth]{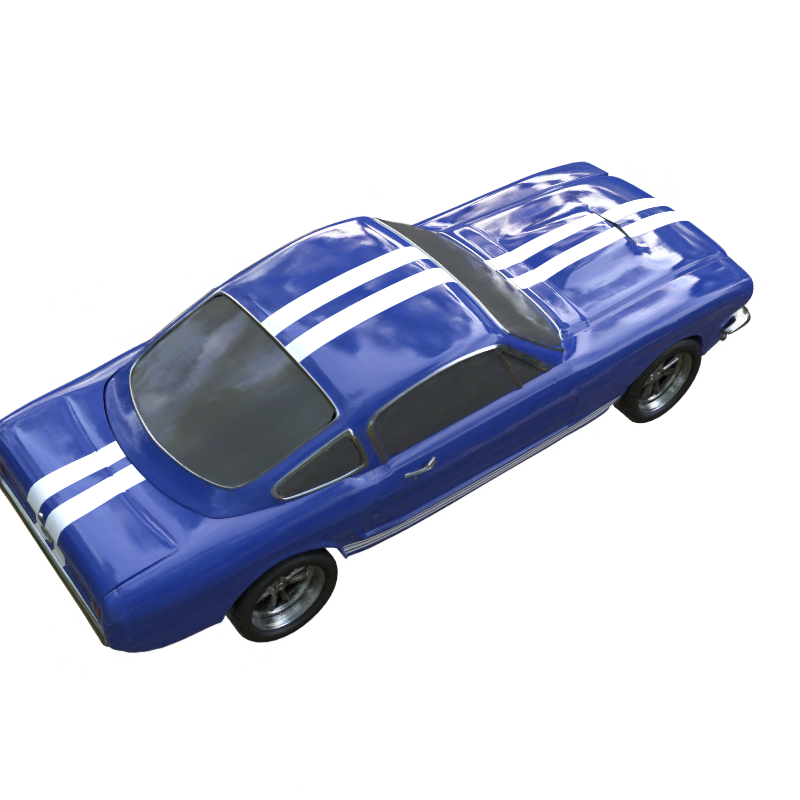}}\put(-20,20){\small{30.82}}\vspace{-0.99cm}\\
\subfloat[GT]{\includegraphics[width=0.14\linewidth]{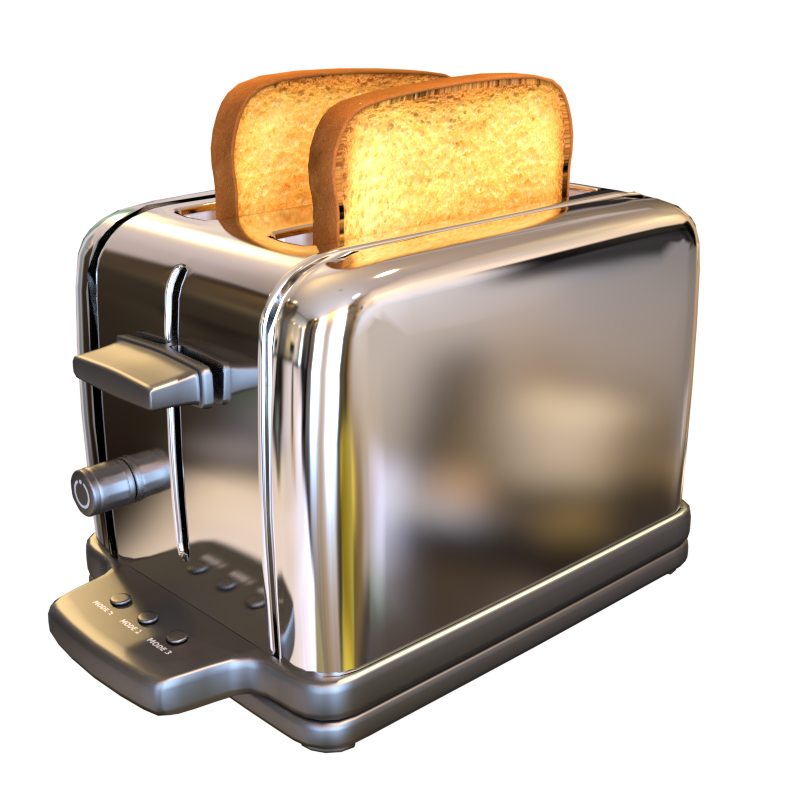}}\hspace*{\fill}
\subfloat[Ours]{\includegraphics[width=0.14\linewidth]
{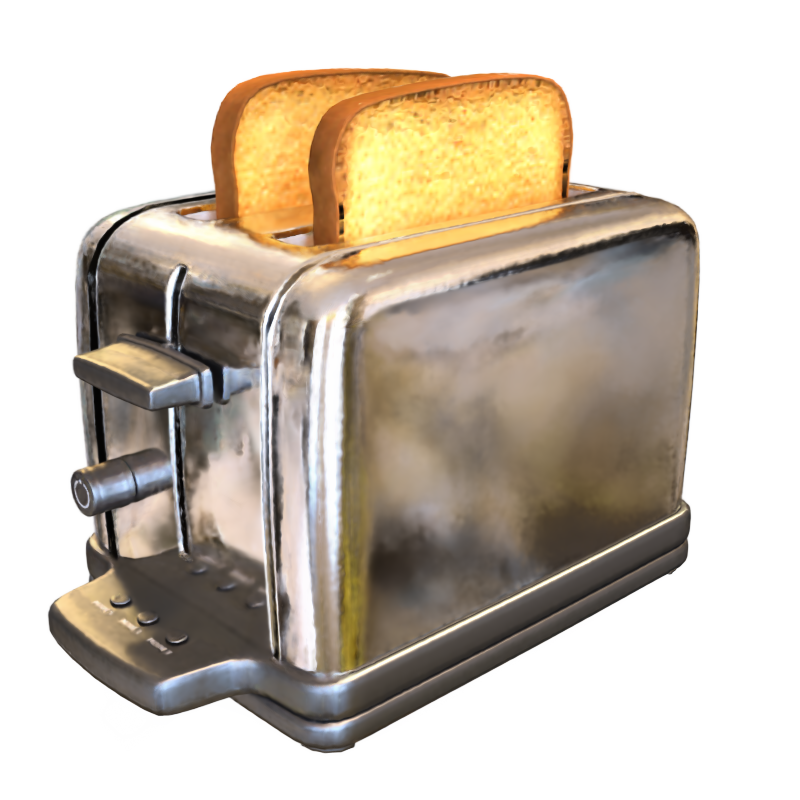}}\put(-20,0){\small{22.91}}\hspace*{\fill}
\subfloat[NeRF]{\includegraphics[width=0.14\linewidth]{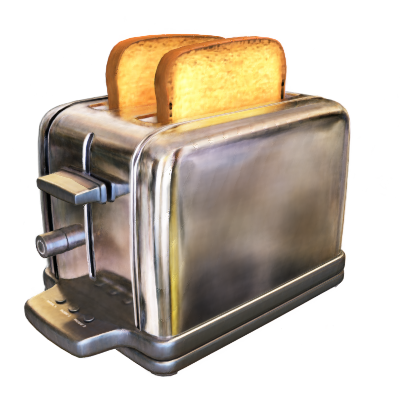}}\put(-20,0){\small{21.72}}\hspace*{\fill}
\subfloat[InstantNGP]{\includegraphics[width=0.14\linewidth]{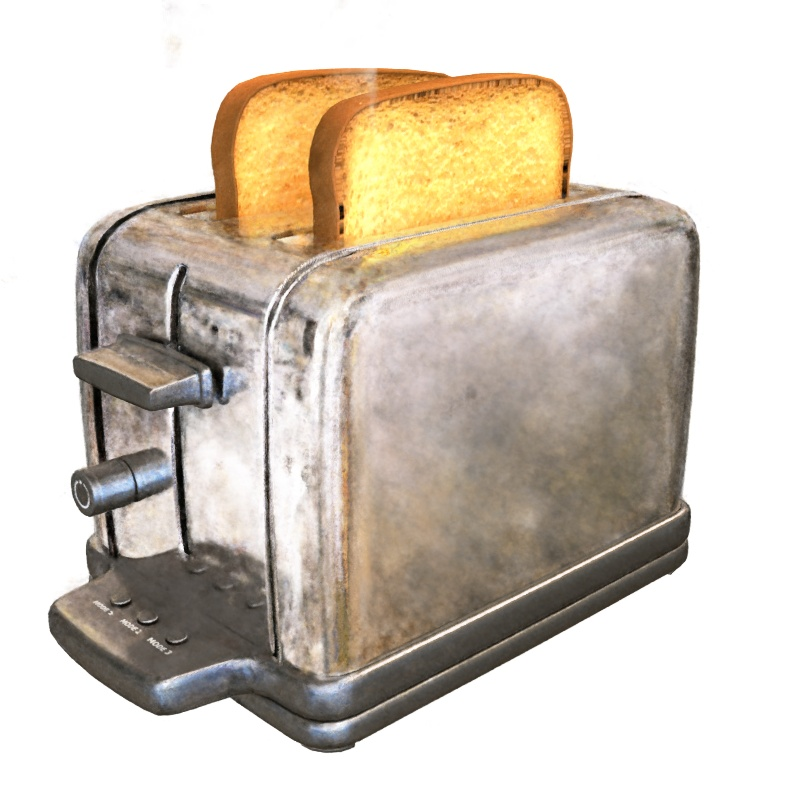}}\put(-20,0){\small{17.2}}\hspace*{\fill}
\subfloat[DVGO]{\includegraphics[width=0.12\linewidth]{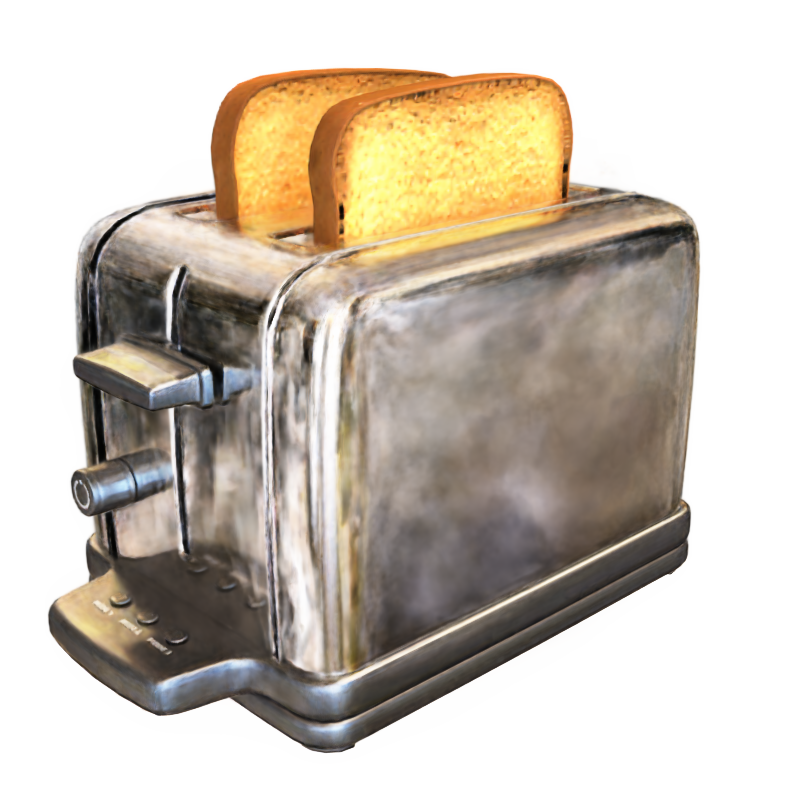}}\put(-20,0){\small{22.31}}\hspace*{\fill}
\subfloat[Ref-NeRF]{\includegraphics[width=0.14\linewidth]{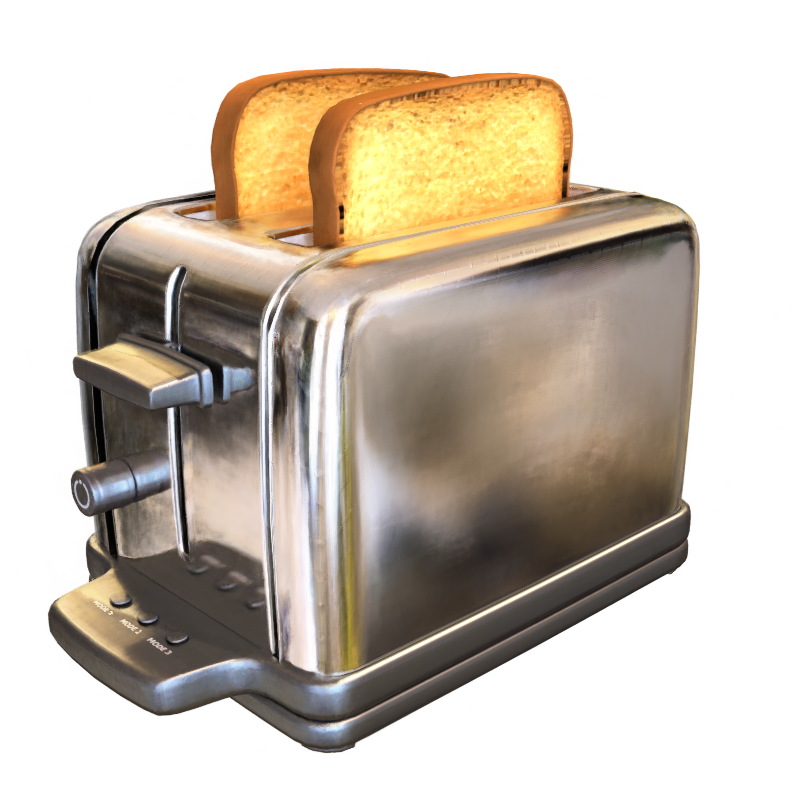}}\put(-20,0){\small{25.70}}
\end{tabular}
\vspace{-0.2cm}
\caption{Qualitative evaluation on two scenes of the synthetic dataset \textit{shiny blender} and the corresponding PSNR scores.}
\label{fig:results:qualitative}
\end{center}
\end{figure*}

\begin{figure}[t]
\begin{center}
\subfloat{\includegraphics[trim=0.3cm 0.2cm 0.2cm 0.2cm,clip=true,width=0.45\linewidth]{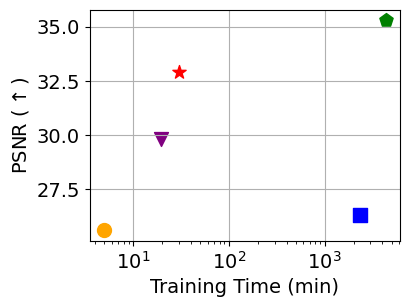}}\hspace{\fill}%
\subfloat{\includegraphics[trim=0.3cm 0.2cm 0.2cm 
0.2cm,clip=true,width=0.45\linewidth]{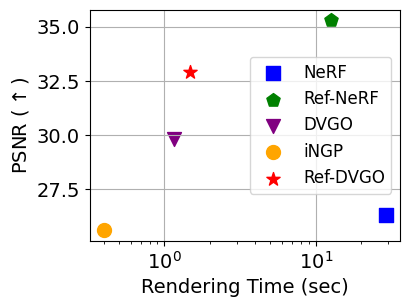}}
\caption{Comparison of our approach against competing methods with regard to PSNR versus training and rendering time on the synthetic dataset \textit{shiny-blender}.}
\label{fig:results:training-rendering-time}
\end{center}
\end{figure}

\noindent\textbf{Results on Shiny Blender Dataset.} In Table \ref{tab:results:shiny-blender} and Fig.~\ref{fig:results:training-rendering-time}, we show the performance of our model against density-based baselines and competing methods on the shiny blender dataset \cite{verbin2022refnerf}. 
%
%
We demonstrate an improved trade-off between reconstruction quality and training/rendering time (note the logarithmic scale for the horizontal time axis) when dealing with such scenes.
The renderings of the examined methods on two out of the six scenes of the shiny blender dataset can be observed in Fig.~\ref{fig:results:qualitative}. \\[-0.2cm]

\begin{table}[t]
\begin{tabular}{l | c c c | c}
\hline
Methods & PSNR $\uparrow$ & SSIM $\uparrow$ & LPIPS $\downarrow$ & Time\\
\hline
\hline
iNGP \cite{mueller2022instant} & 25.61 & 0.900 & 0.143 & \textbf{5m}\\
DVGO \cite{SunSC22dvgo} & {29.83} & 0.933 & 0.135 & \underline{20m}\\
NeRF \cite{mildenhall2020nerf} & 26.31 & 0.935 & - & 2d \\
Ref-NeRf $\cite{verbin2022refnerf}^\dagger$ &  \textbf{35.96} & \textbf{0.967} & \textbf{0.058} & 3d\\
\hline
Ours & \underline{32.89} & \underline{0.957} & \underline{0.104} & {30m} \\
\hline
\end{tabular}
\vspace{-0.1cm}
\caption{Quantitative comparison of our method to competing methods with regard to PSNR, SSIM, LPIPS, and average training time on the six scenes of the shiny-blender dataset \cite{verbin2022refnerf}. We note that the results of InstantNGP \cite{mueller2022instant} were acquired using NerfStudio \cite{nerfstudio} and the symbol $\dagger$ denotes results taken from \cite{verbin2022refnerf}.}
%
%
\label{tab:results:shiny-blender}
\end{table}

\noindent\textbf{Ablation study.} In Table \ref{tab:results:ablation}, we present our ablation study where we demonstrate the importance of the different components of our architecture and optimization process.
As reported, removing the background entropy loss, the per-point RGB loss, the orientation regularization, the total variation regularization, the progressive scaling, the integrated directional encoding, or the reflection direction reparameterization has a negative impact on performance.
Similarly, reducing the number of progressive scaling steps from ten to four or using a smaller directional MLP (3x128 rather than 8x256) has a negative impact.
At the same time, we note that removing the predicted normals penalty marginally improves performance across all metrics. \\[-0.2cm]

\begin{table}[t]
\begin{center}
\begin{tabular}{l | c c c }
\hline
Methods & PSNR $\uparrow$ & SSIM $\uparrow$ & LPIPS $\downarrow$\\
\hline
\hline
Full Model & \underline{32.89} & \underline{0.957} & \underline{0.104} \\
w/o $\mathcal{L}_{bg}$ & 31.15 & 0.957 & 0.105 \\
w/o $\mathcal{L}_{pp}$ & 32.79 & 0.956 & 0.106\\
w/o $\mathcal{R}_{p}$ & \textbf{32.92} & \textbf{0.958} & \textbf{0.103}\\
w/o $\mathcal{R}_{o}$ & 31.13 & 0.947 & 0.119 \\
w/o $TV_{\sigma,c_d,s,b,\rho,\hat{n}'}$ & 32.27 & 0.952 & 0.109 \\
w/o $PGS$ & 27.94 & 0.914 & 0.156 \\
w/o $IDE$ & 29.81 & 0.933 & 0.135 \\
w/o $ref. dir.$ & 30.56 & 0.944 & 0.119\\
w/ $4$ $PGS$ $steps$ & 32.27 & 0.954 & 0.109 \\
w/ small Dir. MLP & 31.92 & 0.949 & 0.118 \\
\hline
\end{tabular}
\vspace{-0.1cm}
\caption{Ablation study of utilized losses, regularizers, and progressive scaling. The results are averaged across all scenes of the shiny blender dataset.}
\label{tab:results:ablation}
\end{center}
\end{table}

\noindent \textbf{Outperforming DVGO.} %
%
%
%
%
%
The color of a point on a highly reflective surface  varies dramatically as the viewing direction changes which makes it hard for the model to find underlying patterns.
Instead it tends to find a solution in a higher dimensional solution space and is thus more susceptible to local minima and overfitting.
The reparameterization constrains the model to a solution space of a lower dimension, which results in more reasonable solutions that are closer to the actual solution in terms of dimensionality.
We evaluate this hypothesis by comparing DVGO and Ref-DVGO on three scenes of the same car with increasing levels of complexity with regard to reflections and specular highlights, as depicted in Table~\ref{tab:results:smart}.
Ref-DVGO appears to be more successful at constraining the solution space in the case of highly reflective scenes and thus outperforms DVGO.\\[-0.2cm]

\begin{table}[t]
\begin{center}
\begin{tabular}{ l | l | c c c }
\hline
Scene & Methods & PSNR $\uparrow$ & SSIM $\uparrow$ & LPIPS $\downarrow$\\
\hline
\hline
\multirow{2}{*}{easy} & DVGO & \textbf{40.2} & \textbf{0.991} & \textbf{0.029} \\
 & Ours & 39.56 & 0.981 & 0.040\\
\hline
\multirow{2}{*}{medium} & DVGO & 31.05 & 0.943 & \textbf{0.086}\\
 & Ours & \textbf{31.34} & \textbf{0.944} & 0.090\\
\hline
\multirow{2}{*}{hard} & DVGO & 29.06 & 0.923 & \textbf{0.098}\\
 & Ours & \textbf{29.40} & \textbf{0.929} & 0.100\\
\hline
\end{tabular}
\end{center}
\vspace{-0.9cm}
\includegraphics[width=0.3\linewidth]{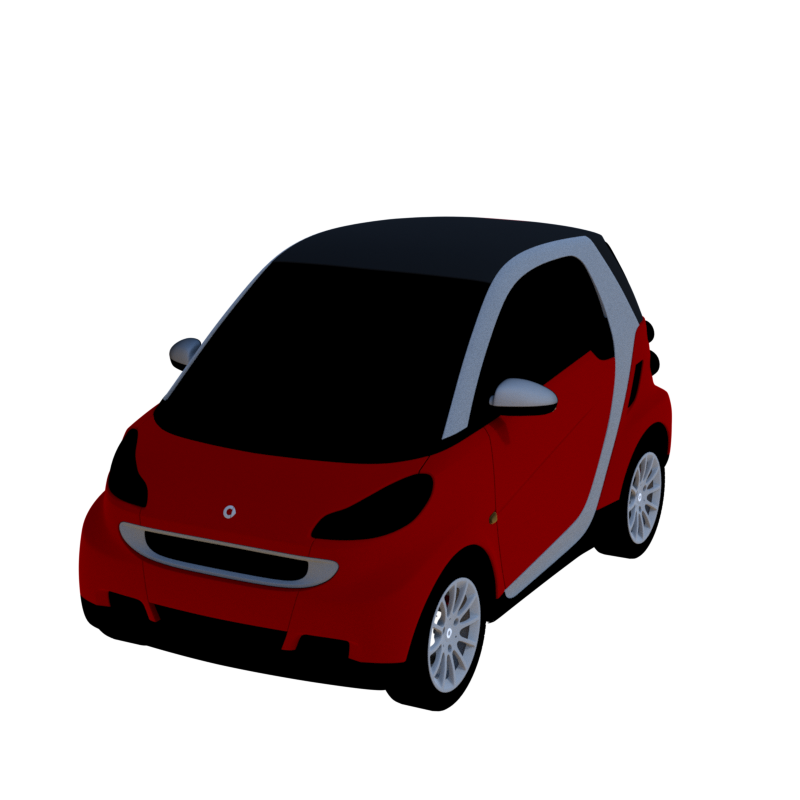}\hspace*{\fill}
\includegraphics[width=0.3\linewidth]{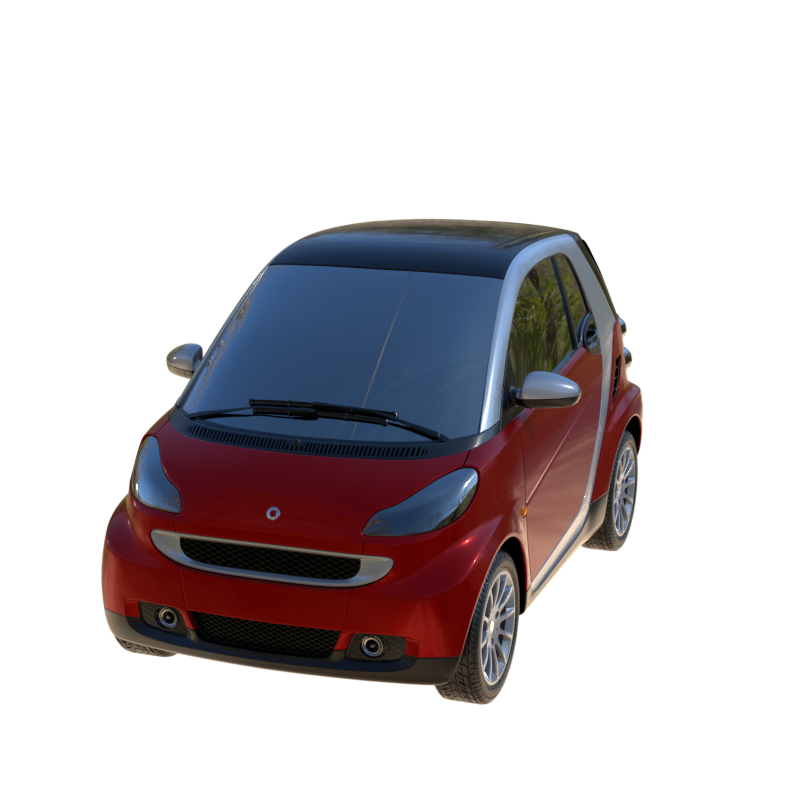}\hspace*{\fill}
\includegraphics[width=0.3\linewidth]{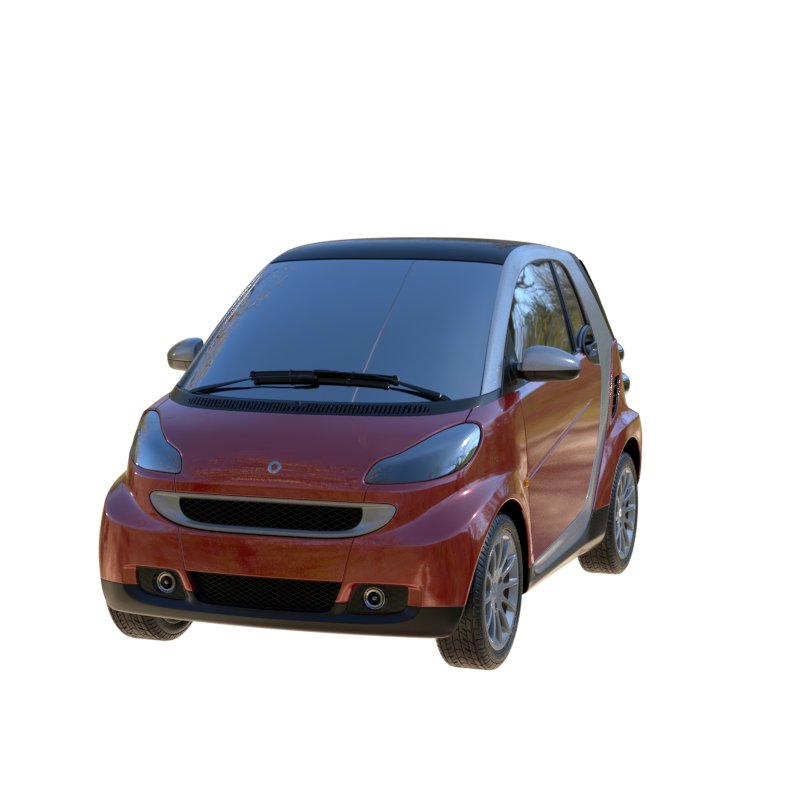}
\vspace{-0.5cm}
\caption{Comparison between DVGO and our method on the smart car scene with increasing levels of reflectivity from easy (left) to medium (middle) and hard (right). For a fairer comparison, we used same-size (8x256) directional MLPs in both methods.}
\label{tab:results:smart}
\end{table}

\noindent \textbf{Performance Gap Against Ref-NeRF.}
Making a hybrid representation reflection-aware improves performance compared to the baselines on reflective scenes, although, bridging the performance gap with Ref-NeRF appears more challenging.
We attribute this gap to the replacement of the spatial MLP with voxel grids that introduce discretization thus breaking spatial continuity and reducing parameter sharing.
This becomes evident when observing how progressive scaling and total variation are essential to learning scenes with complex view-dependent effects and how positively they affect performance by improving spatial continuity.\\[-0.2cm]

\begin{figure}
\subfloat{\includegraphics[trim=7.5cm 4cm 2.5cm 7cm,clip=true,width=0.15\linewidth]{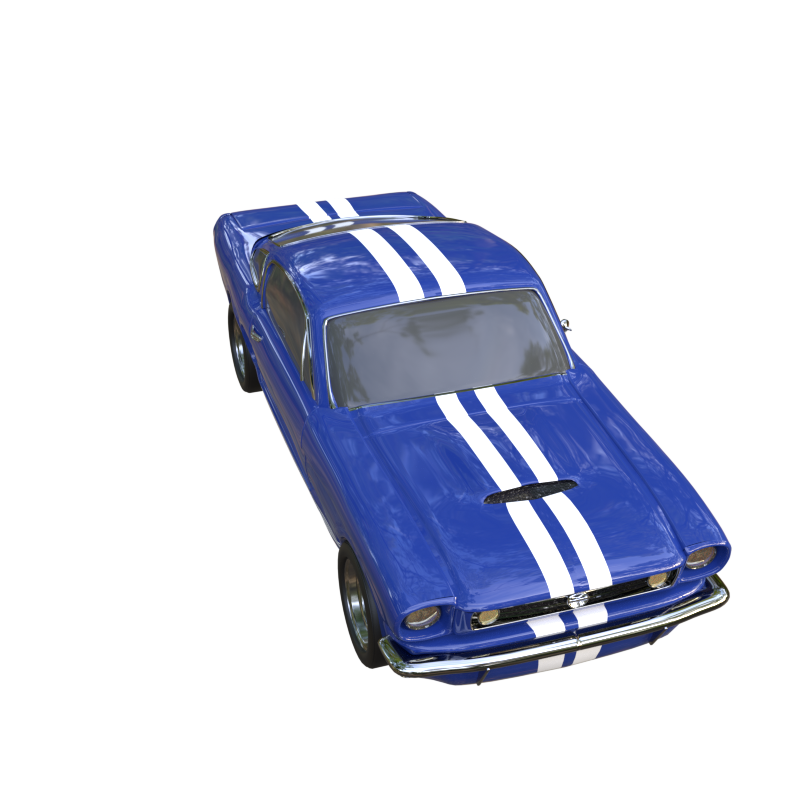}}\hspace{0.1cm}
\subfloat{\includegraphics[trim=7.5cm 4cm 2.5cm 7cm,clip=true,width=0.15\linewidth]{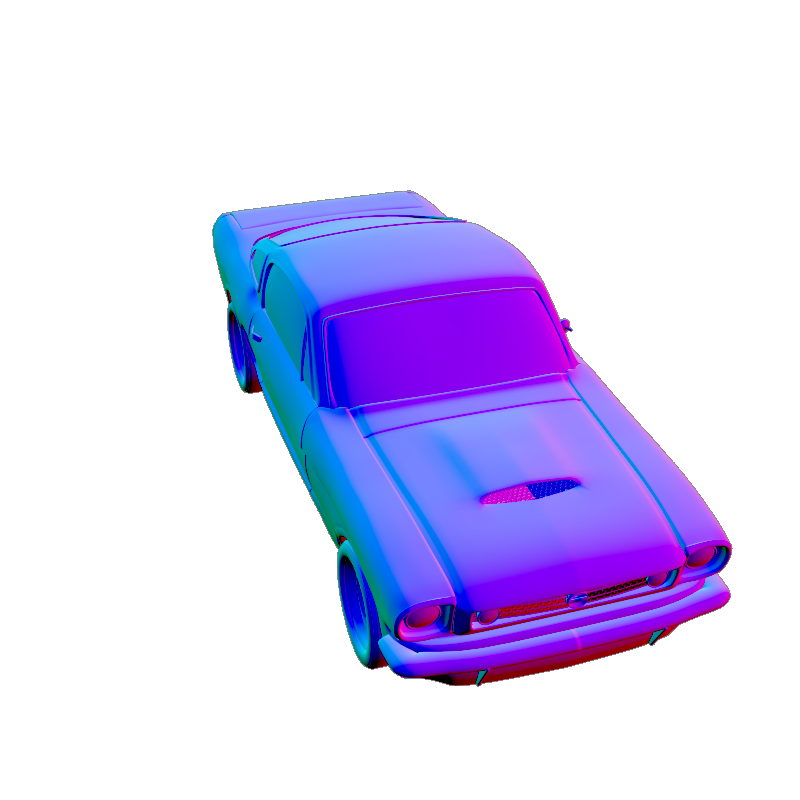}}\hspace{0.1cm}
\subfloat{\includegraphics[trim=7.5cm 4cm 2.5cm 7cm,clip=true,width=0.15\linewidth]{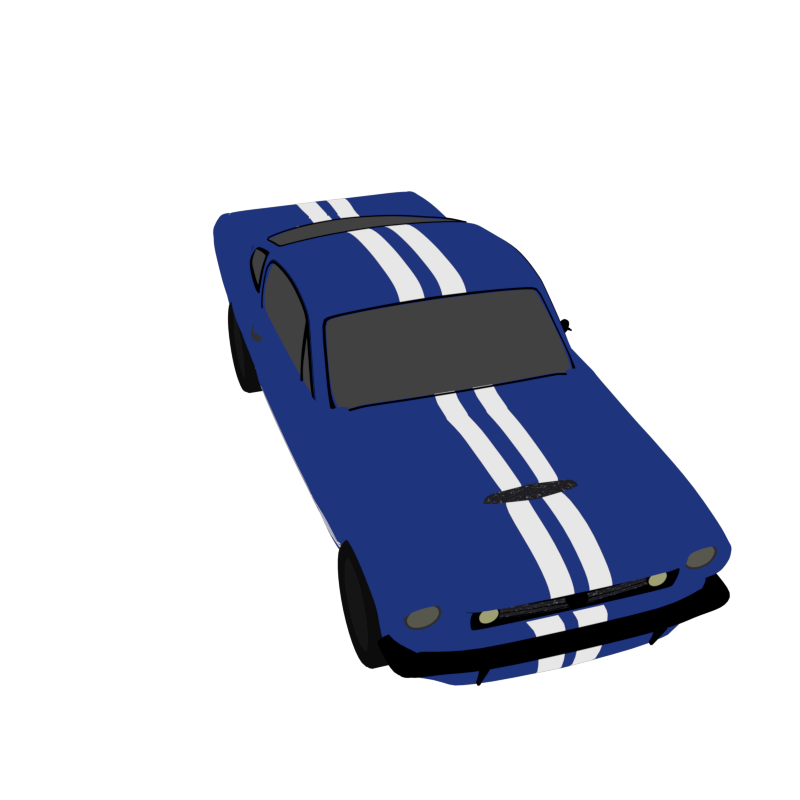}}\hspace{0.1cm}
\subfloat{\includegraphics[trim=7.5cm 4cm 2.5cm 7cm,clip=true,width=0.15\linewidth]{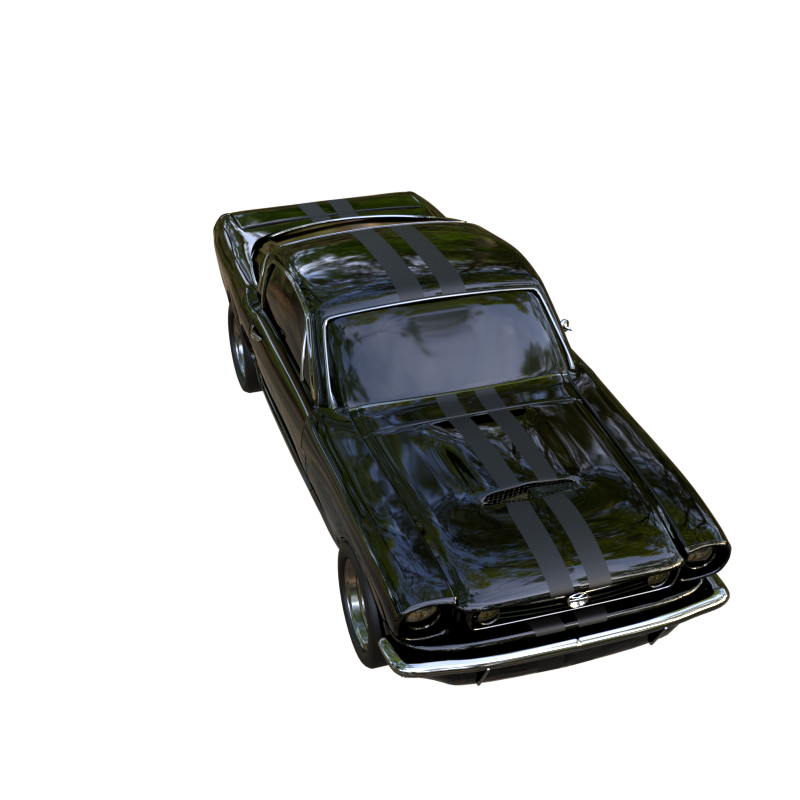}}\\[-0.3cm]
\subfloat{\includegraphics[trim=7.5cm 4cm 2.5cm 7cm,clip=true,width=0.15\linewidth]{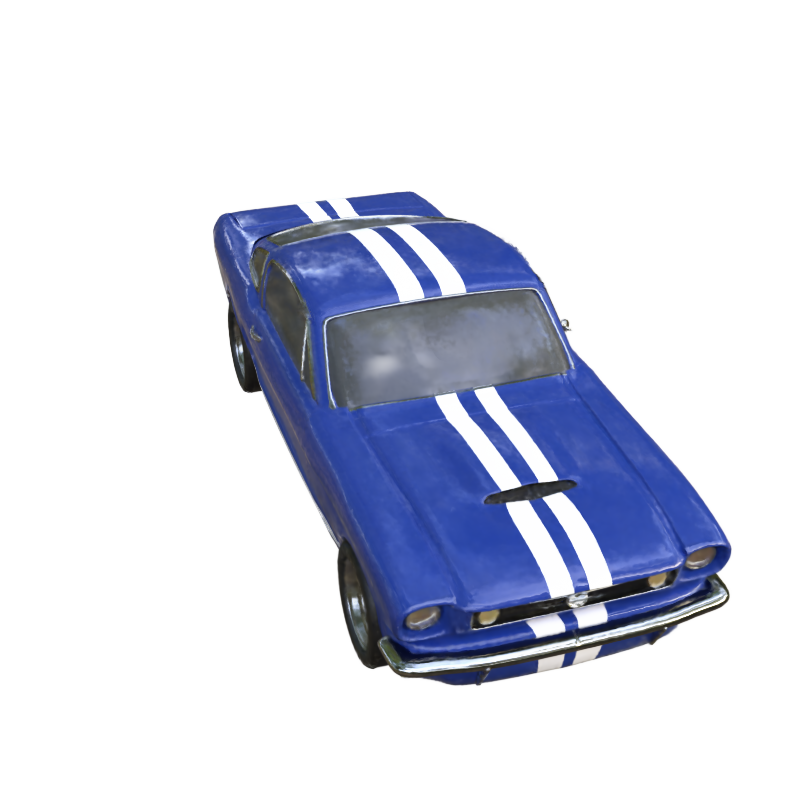}}\hspace*{\fill}
\subfloat{\includegraphics[trim=7.5cm 4cm 2.5cm 7cm,clip=true,width=0.15\linewidth]{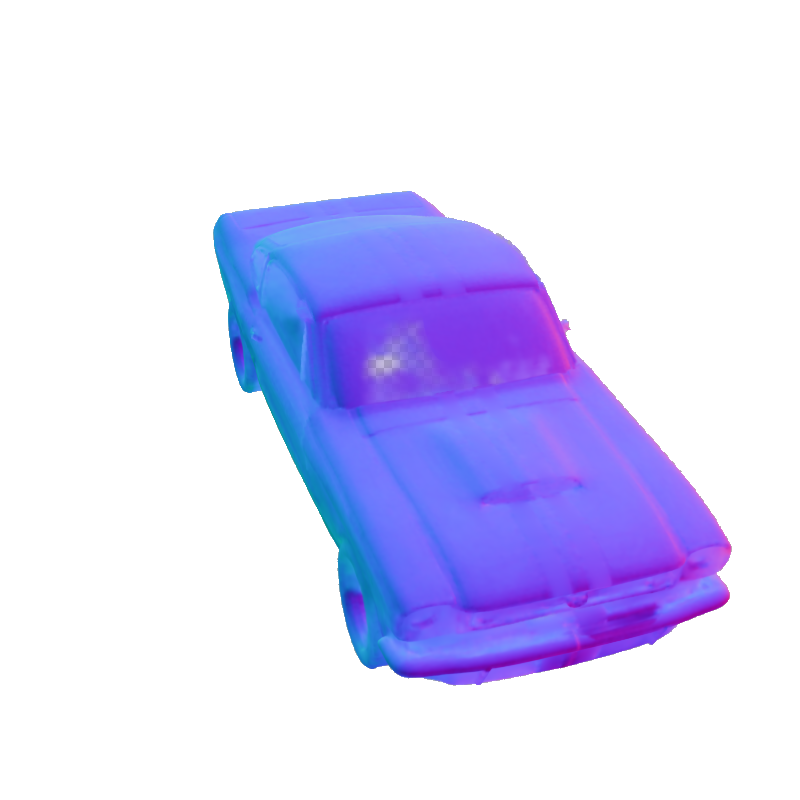}}\hspace*{\fill}
\subfloat{\includegraphics[trim=7.5cm 4cm 2.5cm 7cm,clip=true,width=0.15\linewidth]{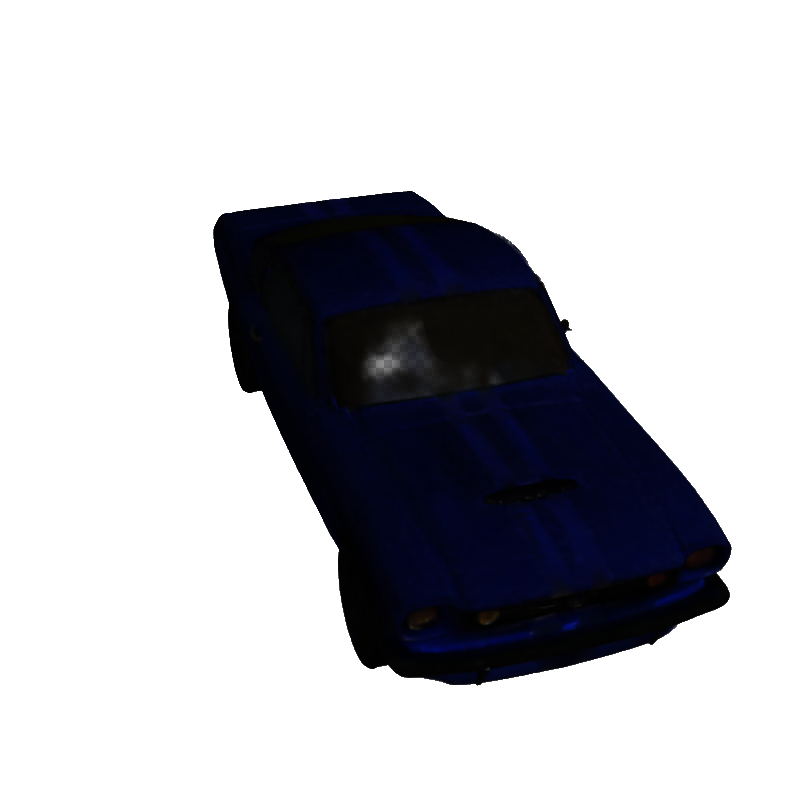}}\hspace*{\fill}
\subfloat{\includegraphics[trim=7.5cm 4cm 2.5cm 7cm,clip=true,width=0.15\linewidth]{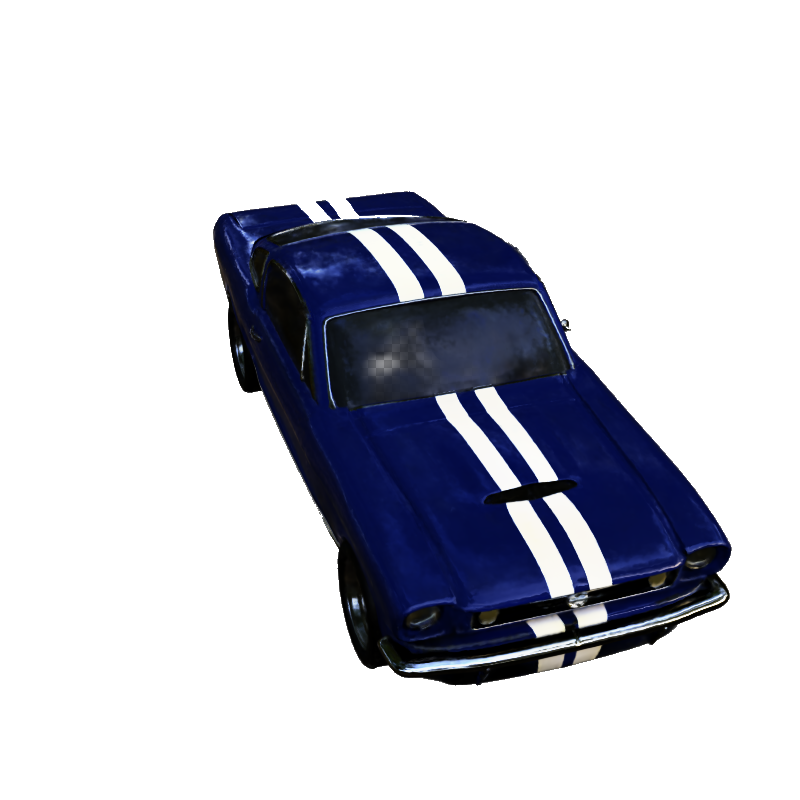}}\hspace*{\fill}
\subfloat{\includegraphics[trim=7.5cm 4cm 2.5cm 7cm,clip=true,width=0.15\linewidth]{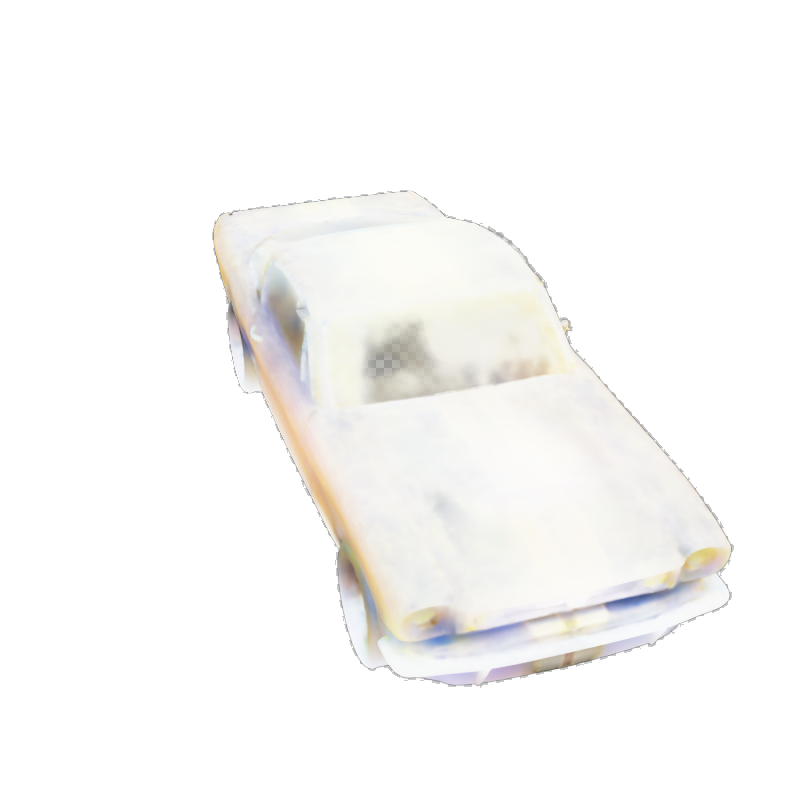}}\hspace*{\fill}
\subfloat{\includegraphics[trim=7.5cm 4cm 2.5cm 7cm,clip=true,width=0.15\linewidth]{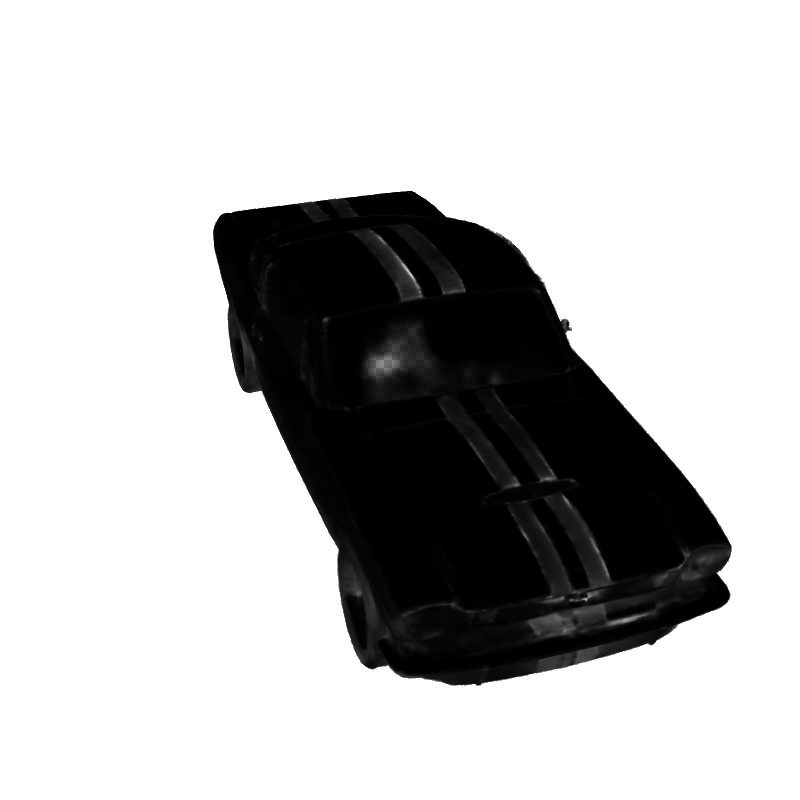}}\\[-0.3cm]
\subfloat[$c$]{\includegraphics[trim=7.5cm 4cm 2.5cm 7cm,clip=true,width=0.15\linewidth]{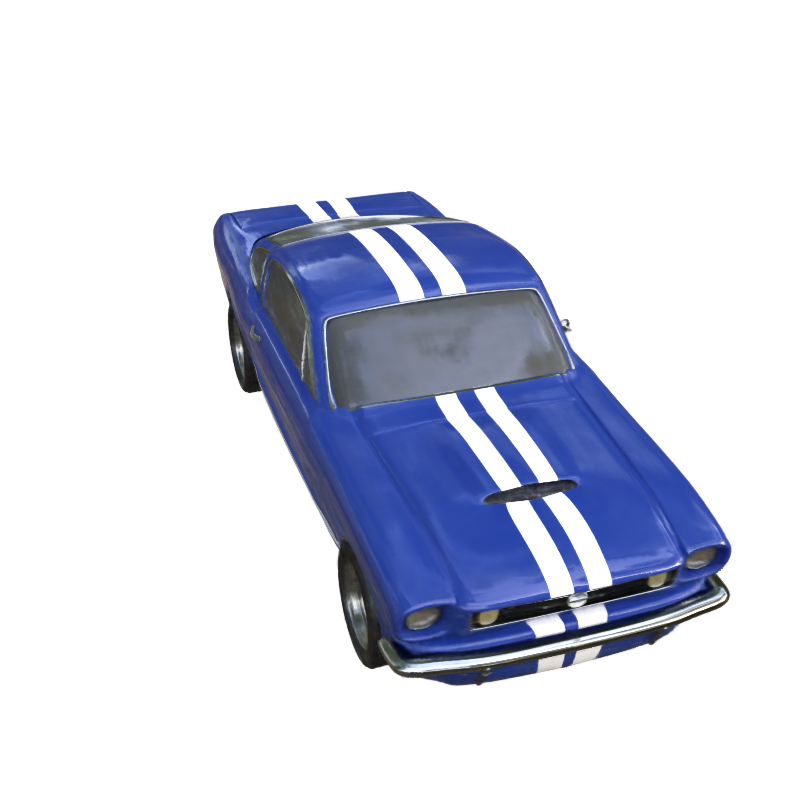}}\hspace*{\fill}
\subfloat[$\hat n'$]{\includegraphics[trim=7.5cm 4cm 2.5cm 7cm,clip=true,width=0.15\linewidth]{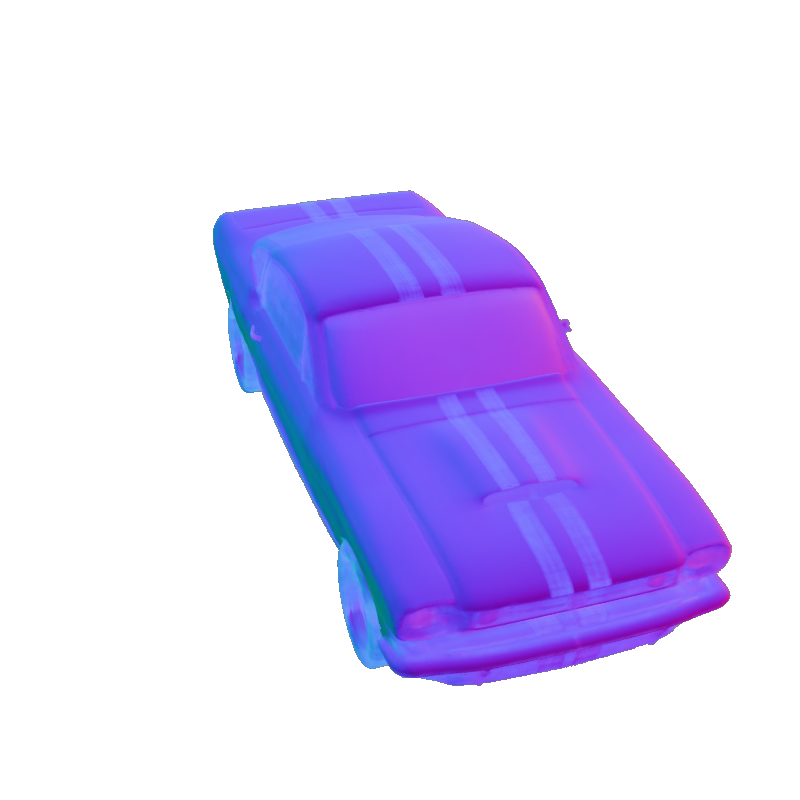}}\hspace*{\fill}
\subfloat[$c_d$]{\includegraphics[trim=7.5cm 4cm 2.5cm 7cm,clip=true,width=0.15\linewidth]{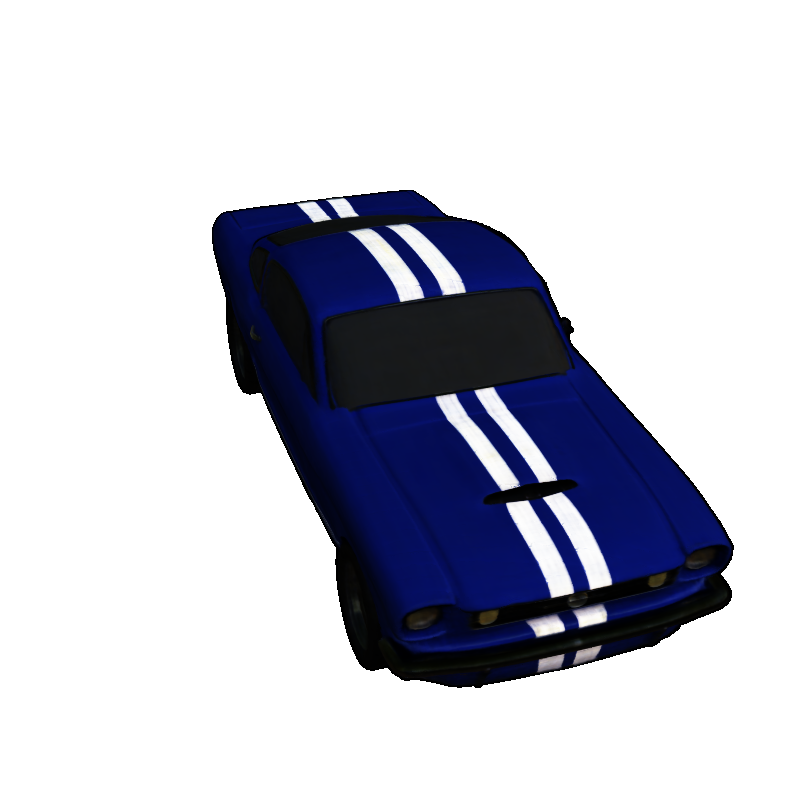}}\hspace*{\fill}
\subfloat[$c_s$]{\includegraphics[trim=7.5cm 4cm 2.5cm 7cm,clip=true,width=0.15\linewidth]{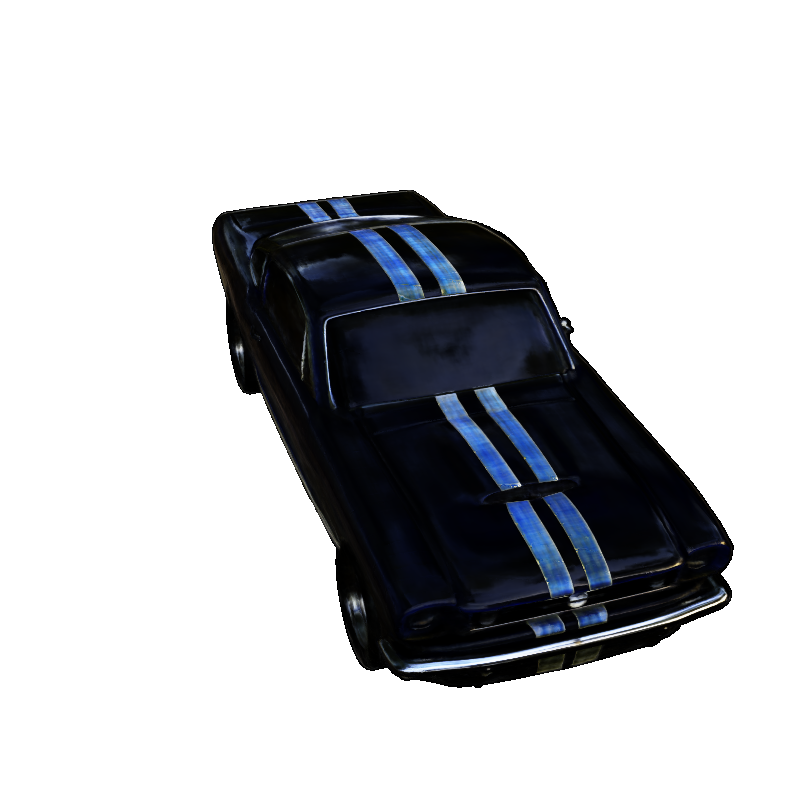}}\hspace*{\fill}
\subfloat[$s$]{\includegraphics[trim=7.5cm 4cm 2.5cm 7cm,clip=true,width=0.15\linewidth]{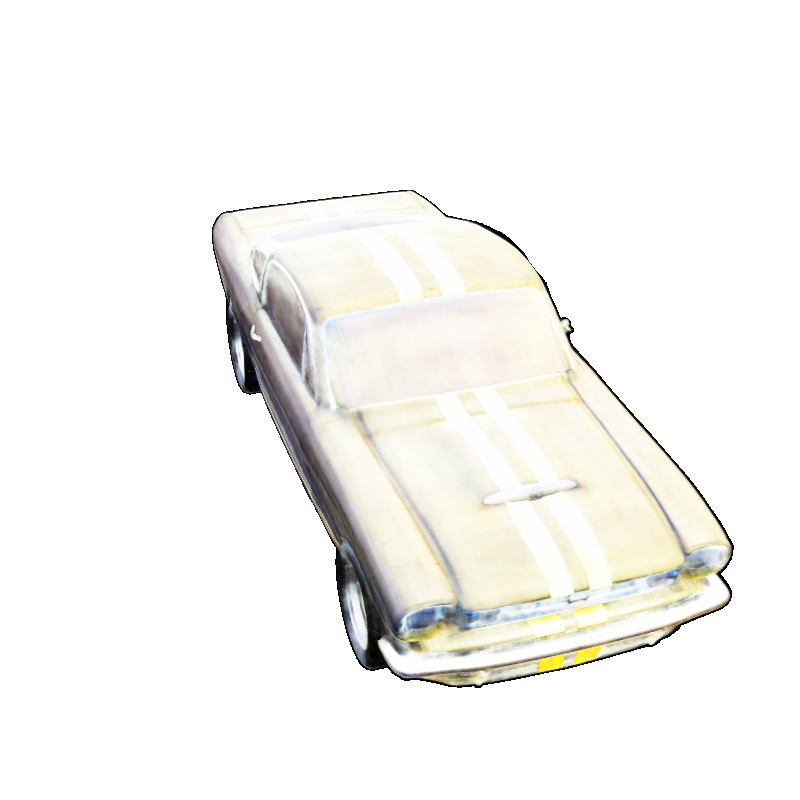}}\hspace*{\fill}
\subfloat[$\rho$]{\includegraphics[trim=7.5cm 4cm 2.5cm 7cm,clip=true,width=0.15\linewidth]{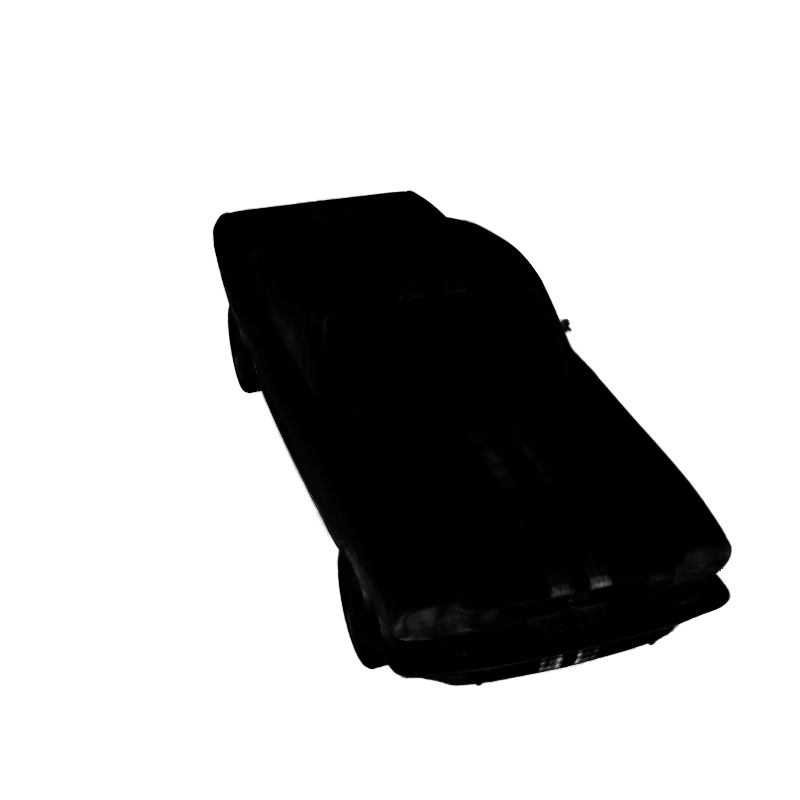}}\\
\vspace{-0.4cm}
\caption{Qualitative comparison of rendered images and scene properties between ground truth (top), our method (middle), and Ref-NeRF (bottom).}
\label{fig:results:materials}
\end{figure}

\noindent\textbf{Limitations.} Our proposed reflection-aware hybrid representation tends to perform better  than the baselines on reflective scenes, however, there is still a large performance gap compared to Ref-NeRF. 
Both qualitative and quantitative results demonstrate that Ref-NeRF is superior, although, at the expense of training and rendering time.
Both methods are susceptible to estimating material properties that are not physically plausible even though they result in plausible renderings.
According to Fig.~\ref{fig:results:materials}, the hybrid representation, however, appears to have even greater difficulty in disentangling view-consistent and view-dependent appearance in favor of the latter, while also resulting in more semi-transparent surfaces and artifacts i.e. holes in the windshield.
%
Nevertheless, our method resulted in a better disentanglement of the white stripes from the geometry of the car, as depicted in the normal renderings.

\section{Conclusion}

In this work, we investigate the feasibility of improving and accelerating neural rendering for reflective scenes through a hybrid implicit-explicit representation. 
To this end, we adapt a fully implicit reflection-aware neural rendering model into a hybrid implicit-explicit representation. 
Through our evaluation, we demonstrate an improved speed-accuracy trade-off on scenes with reflective objects compared to previous non-reflection-aware methods,
At the same time, we discuss our hypotheses about the reasons behind the increase in performance and the failure to reach the rendering quality of Ref-NeRF due to the inherent spatial discretization of hybrid methods.

\section*{Acknowledgements}
We gratefully acknowledge funding support from the Sim2Real2 project, in the context of the Ford-KU Leuven alliance program.

{\small
\bibliographystyle{ieee_fullname}
\bibliography{egpaper}

\begin{thebibliography}{10}\itemsep=-1pt

\bibitem{barron2021mipnerf}
Jonathan~T. Barron, Ben Mildenhall, Matthew Tancik, Peter Hedman, Ricardo
  Martin-Brualla, and Pratul~P. Srinivasan.
\newblock Mip-nerf: A multiscale representation for anti-aliasing neural
  radiance fields, 2021.

\bibitem{barron2022mipnerf360}
Jonathan~T. Barron, Ben Mildenhall, Dor Verbin, Pratul~P. Srinivasan, and Peter
  Hedman.
\newblock Mip-nerf 360: Unbounded anti-aliased neural radiance fields.
\newblock {\em CVPR}, 2022.

\bibitem{Chen2022ECCV}
Anpei Chen, Zexiang Xu, Andreas Geiger, Jingyi Yu, and Hao Su.
\newblock Tensorf: Tensorial radiance fields.
\newblock In {\em European Conference on Computer Vision (ECCV)}, 2022.

\bibitem{fan2023factoredneus}
Yue Fan, Ivan Skorokhodov, Oleg Voynov, Savva Ignatyev, Evgeny Burnaev, Peter
  Wonka, and Yiqun Wang.
\newblock Factored-neus: Reconstructing surfaces, illumination, and materials
  of possibly glossy objects, 2023.

\bibitem{ge2023ref}
Wenhang Ge, Tao Hu, Haoyu Zhao, Shu Liu, and Ying-Cong Chen.
\newblock Ref-neus: Ambiguity-reduced neural implicit surface learning for
  multi-view reconstruction with reflection.
\newblock {\em arXiv preprint arXiv:2303.10840}, 2023.

\bibitem{Liang2023ENVIDRID}
Ruofan Liang, Hui-Hsia Chen, Chunlin Li, Fan Chen, Selvakumar Panneer, and
  Nandita Vijaykumar.
\newblock Envidr: Implicit differentiable renderer with neural environment
  lighting.
\newblock {\em ArXiv}, abs/2303.13022, 2023.

\bibitem{mildenhall2020nerf}
Ben Mildenhall, Pratul~P. Srinivasan, Matthew Tancik, Jonathan~T. Barron, Ravi
  Ramamoorthi, and Ren Ng.
\newblock Nerf: Representing scenes as neural radiance fields for view
  synthesis.
\newblock In {\em ECCV}, 2020.

\bibitem{mueller2022instant}
Thomas M\"uller, Alex Evans, Christoph Schied, and Alexander Keller.
\newblock Instant neural graphics primitives with a multiresolution hash
  encoding.
\newblock {\em ACM Trans. Graph.}, 41(4):102:1--102:15, July 2022.

\bibitem{roessle2022depthpriorsnerf}
Barbara Roessle, Jonathan~T. Barron, Ben Mildenhall, Pratul~P. Srinivasan, and
  Matthias Nie{\ss}ner.
\newblock Dense depth priors for neural radiance fields from sparse input
  views.
\newblock In {\em Proceedings of the IEEE/CVF Conference on Computer Vision and
  Pattern Recognition (CVPR)}, June 2022.

\bibitem{rudin1994tv}
L.I. Rudin and S. Osher.
\newblock Total variation based image restoration with free local constraints.
\newblock In {\em Proceedings of 1st International Conference on Image
  Processing}, volume~1, pages 31--35 vol.1, 1994.

\bibitem{yu_and_fridovichkeil2021plenoxels}
{Sara Fridovich-Keil and Alex Yu}, Matthew Tancik, Qinhong Chen, Benjamin
  Recht, and Angjoo Kanazawa.
\newblock Plenoxels: Radiance fields without neural networks.
\newblock In {\em CVPR}, 2022.

\bibitem{SunSC22dvgo}
Cheng Sun, Min Sun, and Hwann{-}Tzong Chen.
\newblock Direct voxel grid optimization: Super-fast convergence for radiance
  fields reconstruction.
\newblock In {\em CVPR}, 2022.

\bibitem{sun2022improved}
Cheng Sun, Min Sun, and Hwann-Tzong Chen.
\newblock Improved direct voxel grid optimization for radiance fields
  reconstruction, 2022.

\bibitem{nerfstudio}
Matthew Tancik, Ethan Weber, Evonne Ng, Ruilong Li, Brent Yi, Justin Kerr,
  Terrance Wang, Alexander Kristoffersen, Jake Austin, Kamyar Salahi, Abhik
  Ahuja, David McAllister, and Angjoo Kanazawa.
\newblock Nerfstudio: A modular framework for neural radiance field
  development.
\newblock In {\em ACM SIGGRAPH 2023 Conference Proceedings}, SIGGRAPH '23,
  2023.

\bibitem{verbin2022refnerf}
Dor Verbin, Peter Hedman, Ben Mildenhall, Todd Zickler, Jonathan~T. Barron, and
  Pratul~P. Srinivasan.
\newblock {Ref-NeRF}: Structured view-dependent appearance for neural radiance
  fields.
\newblock {\em CVPR}, 2022.

\bibitem{yu2020pixelnerf}
Alex Yu, Vickie Ye, Matthew Tancik, and Angjoo Kanazawa.
\newblock pixelnerf: Neural radiance fields from one or few images.
\newblock In {\em CVPR}, 2021.

\end{thebibliography}
}

\end{document}